\documentclass[sigconf,nonacm=true]{acmart}

\usepackage{hyperref}
\usepackage{hyperxmp}
\usepackage{hyperxmp}
\usepackage{booktabs}
\usepackage{microtype}
\usepackage{graphicx}
\usepackage{tikz}
\usetikzlibrary{arrows.meta,positioning,fit,backgrounds,calc}

\newcommand{\NScenarios}{40}
\newcommand{\NDyads}{160}
\newcommand{\NProbes}{2880}
\newcommand{\MeanTurns}{5.94}
\newcommand{\ParseFailures}{0}
\newcommand{\AgreementRate}{96.2}
\newcommand{\ParetoRate}{0.7}
\newcommand{\EfficiencyLoss}{20.5}

\newcommand{\CompatMissRate}{76.6}

\newcommand{\EffHonest}{19.5}

\newcommand{\EffStrategic}{21.6}

\newcommand{\TopAnalystBase}{53.8}
\newcommand{\TauAnalystBase}{0.501}
\newcommand{\CmpAnalystBase}{35.0}
\newcommand{\ResAnalystBase}{22.7}

\newcommand{\TopAnalystInverse}{51.6}
\newcommand{\TauAnalystInverse}{0.435}
\newcommand{\CmpAnalystInverse}{35.3}
\newcommand{\ResAnalystInverse}{22.3}

\newcommand{\TopAnalystMirror}{66.6}
\newcommand{\TauAnalystMirror}{0.554}
\newcommand{\CmpAnalystMirror}{34.1}
\newcommand{\ResAnalystMirror}{22.2}

\newcommand{\TopLabelJunior}{52.8}
\newcommand{\TauLabelJunior}{0.512}
\newcommand{\CmpLabelJunior}{34.1}
\newcommand{\ResLabelJunior}{22.3}

\newcommand{\TopLabelSenior}{52.8}
\newcommand{\TauLabelSenior}{0.491}
\newcommand{\CmpLabelSenior}{34.1}
\newcommand{\ResLabelSenior}{23.4}

\newcommand{\TopParticipantBase}{51.2}
\newcommand{\TauParticipantBase}{0.491}
\newcommand{\CmpParticipantBase}{36.9}
\newcommand{\ResParticipantBase}{22.9}

\newcommand{\TopToneAccommodating}{47.2}
\newcommand{\TauToneAccommodating}{0.479}
\newcommand{\CmpToneAccommodating}{39.4}
\newcommand{\ResToneAccommodating}{22.8}

\newcommand{\TopToneAssertive}{52.5}
\newcommand{\TauToneAssertive}{0.502}
\newcommand{\CmpToneAssertive}{36.2}
\newcommand{\ResToneAssertive}{23.3}

\newcommand{\ProjShift}{6.5}
\newcommand{\ProjShiftCI}{[6.0, 7.0]}
\newcommand{\ToneTopMean}{5.3}
\newcommand{\ToneTopCI}{[1.9, 9.1]}

\newcommand{\ToneTauMean}{0.024}
\newcommand{\ToneTauCI}{[0.002, 0.047]}

\newcommand{\ToneResLevelMean}{0.43}
\newcommand{\ToneResLevelCI}{[-0.06, 0.87]}

\newcommand{\IdTopMean}{0.0}
\newcommand{\IdTopCI}{[-2.8, 2.8]}

\newcommand{\IdTauMean}{-0.021}
\newcommand{\IdTauCI}{[-0.038, -0.005]}

\newcommand{\IdResMean}{1.09}
\newcommand{\IdResCI}{[0.77, 1.43]}

\newcommand{\IdResLevelMean}{1.10}
\newcommand{\IdResLevelCI}{[0.78, 1.44]}

\newcommand{\SecondOrderHit}{72.5}
\newcommand{\AssumedTransparency}{68.8}
\newcommand{\PartnerActuallyRight}{51.2}
\newcommand{\TransGapMean}{-3.8}
\newcommand{\TransGapCI}{[-10.9, 3.4]}

\newcommand{\FrameTopMean}{-2.5}
\newcommand{\FrameTopCI}{[-5.3, 0.3]}

\newcommand{\ProjTopMean}{15.0}
\newcommand{\ProjTopCI}{[10.9, 19.1]}

\newcommand{\ProjTopCellMin}{13.8}
\newcommand{\ProjTopCellMax}{17.5}

\newcommand{\ProjTopCells}{4}
\newcommand{\ProjTauMean}{0.119}
\newcommand{\ProjTauCI}{[0.096, 0.144]}

\newcommand{\BaseParetoRate}{1.1}
\newcommand{\BaseEfficiencyLoss}{24.1}
\newcommand{\BaseCompatHitRate}{24.8}

\newcommand{\ParetoGapMean}{-0.4}
\newcommand{\ParetoGapCI}{[-1.1, 0.9]}

\newcommand{\EffGapMean}{-3.7}
\newcommand{\EffGapCI}{[-5.1, -2.4]}

\newcommand{\CompatGapMean}{-1.5}
\newcommand{\CompatGapCI}{[-8.0, 5.6]}

\newcommand{\HonEffMean}{2.1}
\newcommand{\HonEffCI}{[-0.8, 4.8]}

\newcommand{\ConfMin}{0.35}
\newcommand{\ConfMax}{0.65}
\newcommand{\ConfCorrect}{0.537}
\newcommand{\ConfIncorrect}{0.520}
\newcommand{\ConfGapMean}{0.018}
\newcommand{\ConfGapCI}{[0.013, 0.022]}
\newcommand{\ConfGapN}{2560}
\newcommand{\LinkN}{154}

\newcommand{\LinkEffHigh}{20.0}
\newcommand{\LinkEffLow}{21.3}
\newcommand{\LinkCompatHigh}{26.7}
\newcommand{\LinkCompatLow}{18.8}

\setcopyright{none}
\renewcommand\footnotetextcopyrightpermission[1]{}
\acmYear{2026}
\copyrightyear{2026}

\begin{document}

\title{Mind or Message? Auditing Theory of Mind in \\Multi-Agent Social Simulation}

\author{Cong Li}
\affiliation{\institution{DailyAdvance}\country{}}
\author{Cheng Chen}
\affiliation{\institution{DailyAdvance}\country{}}
\author{Thomas Fung}
\affiliation{\institution{DailyAdvance}\country{}}
\author{Alex Rossi}
\affiliation{\institution{DailyAdvance}\country{}}
\author{Yi Li}
\affiliation{\institution{DailyAdvance}\country{}}

\begin{abstract}
Language model agents are increasingly used to simulate social interaction, and the resulting transcripts read as though the agents understand one another. We ask whether that appearance rests on a model of the partner's mind or on the surface record of what the partner said. We build a social simulation in which both questions have exact answers: \NScenarios\ multi-issue negotiations whose hidden preference weights and whose full Pareto frontier are known by construction. Two model families negotiate across \NDyads\ dyads, every transcript is frozen before any measurement, and \NProbes\ counterfactual probes then hold the evidence byte identical while moving one factor at a time: the reader's own stake, the partner's tone, an identity label, and the order of recursion. The agents are socially fluent and economically poor. They reach agreement in \AgreementRate\% of dyads with \ParseFailures\ protocol failures, yet only \ParetoRate\% of deals land on the Pareto frontier, they leave \EfficiencyLoss\% of the available joint value unclaimed, and they miss the one issue on which their interests are perfectly aligned in \CompatMissRate\% of deals; on the frontier and on that aligned issue, a package drawn at random from the set both sides would accept does as well. The probes locate the failure. Swapping only the reader's own payoff sheet, while the partner's words and offers stay identical, moves the inferred top priority by \ProjTopMean\ percentage points, which is egocentric projection rather than inference, while a tone rewrite moves it by \ToneTopMean\ percentage points and an identity label by \IdTopMean. Most tellingly, an agent predicts what its partner believes about it \SecondOrderHit\% of the time while that partner's belief is itself correct only \PartnerActuallyRight\% of the time: the agents track the conversation far better than they track the mind behind it. We argue that social simulations should be validated against latent ground truth rather than surface plausibility, and we release a protocol that makes such validation mechanical.

\end{abstract}



\maketitle
\raggedbottom

\section{Introduction}
\label{sec:introduction}

Social simulation with language model agents has moved quickly from demonstration to method. Agents populate towns, staff focus groups, stand in for survey respondents, and rehearse conversations that researchers then read as evidence about how people might behave \cite{park2023generative,park2022social,argyle2023one,aher2023using}. The transcripts are strikingly readable: agents take turns, acknowledge each other's concerns, concede, and close. Because the output looks like a social interaction, it is tempting to treat it as one, and to treat whatever regularity appears in the transcript as a finding about social behavior rather than about text generation.

That inference has a load bearing assumption underneath it. A social interaction is not only an exchange of sentences; it is an exchange in which each party maintains some representation of what the other wants, knows, and will accept, and adjusts accordingly. This representation is what the theory of mind literature calls a partner model \cite{premack1978chimpanzee,wimmer1983beliefs}. If simulated agents maintain partner models that track the hidden states they are supposed to track, the simulation carries information about social processes. If instead they produce locally plausible continuations without such a representation, the transcript can look cooperative while carrying no information about the mechanism a researcher wants to study. The two cases are indistinguishable from the transcript alone, which is exactly what most social simulation work reads.

Testing the distinction requires a setting where the mental state has a known answer. Existing theory of mind evaluations largely use story comprehension items with a labeled correct inference \cite{sap2022neural,kim2023fantom,gandhi2023understanding,kosinski2024evaluating}, and they have shown that performance is fragile to superficial rewording \cite{ullman2023trivial,shapira2024clever}. They are also static: nothing the model says changes what it is asked to infer, and no social outcome depends on getting the inference right. Social simulation has the opposite profile. It is genuinely interactive, but the partner's mental state is usually whatever the prompt said it was, with no independent way to check whether the other agent recovered it. We want both properties at once: a live interaction in which the hidden state is fixed in advance and the quality of the interaction can be scored against it.

Multi-issue integrative negotiation supplies exactly this \cite{raiffa1982art,bazerman2000negotiation}. Each side holds a private vector of preference weights over several issues. Some issues are genuinely opposed, some can be traded because the sides weight them differently, and at least one is compatible, meaning both sides want the same outcome and merely have to notice. This gives two exact ground truths from the same generator: the partner's hidden weight vector, top priority, and walk away value, and the full Pareto frontier, which we compute by enumeration so that every agreement can be scored for how much joint value it left behind. Human negotiators fail here in a specific way, systematically assuming conflict where interests are aligned, a robust effect called the incompatibility bias \cite{thompson1990social,thompson1991information}. The paradigm therefore also tells us what a human like failure would look like.

We use this to run a training free audit. Two model families negotiate across \NDyads\ dyads drawn from \NScenarios\ scenarios, under an honest and a strategic disclosure condition. Every transcript is frozen before any measurement. We then apply \NProbes\ counterfactual probes to the frozen record, each of which holds the evidence byte identical and changes exactly one thing: the reader's own payoff sheet, the partner's tone with the offer sequence preserved, an identity label attached to the partner, or the direction of recursion. Because the evidence cannot move, any change in the reported partner model is attributable to the manipulated factor rather than to new information. This is the design that separates a model of the mind from a reading of the message.

The results separate the two cleanly, and not in the direction the transcripts suggest. The agents are socially fluent, reaching agreement in \AgreementRate\% of dyads with \ParseFailures\ protocol failures, and economically poor: only \ParetoRate\% of agreements reach the Pareto frontier, \EfficiencyLoss\% of available joint value is destroyed, and the compatible issue, where both sides want the same thing, is missed in \CompatMissRate\% of deals. Against an exact baseline, a random package from the set both sides would accept, the agents recover some joint value but are indistinguishable on the other two counts. The probes show why. Swapping only the reader's own payoff sheet, with the partner's words and offers unchanged, shifts the inferred top priority by \ProjTopMean\ percentage points, which is projection rather than inference, while a tone rewrite moves it by \ToneTopMean\ percentage points and an identity label by \IdTopMean. And an agent predicts its partner's stated belief about it \SecondOrderHit\% of the time while that belief is correct only \PartnerActuallyRight\% of the time, so the agents track the shared conversational record far better than the hidden state it is supposed to reveal.

This paper contributes a negotiation based audit protocol in which both the mental state and the social outcome have exact answers, paired counterfactual probes that isolate projection, tone, identity, and recursion on frozen evidence, and an empirical account of how simulated agents can be fluent at the surface while failing to recover the hidden states that would make the simulation informative. Our aim is diagnostic. We do not argue that social simulation is invalid. We argue that its validity is an empirical question that needs latent ground truth to answer, and we provide one way to supply it.

\section{Related Work}
\label{sec:related}

\paragraph{Theory of mind and simulation validity.}
Whether language models represent other minds has been contested since large models began passing textual false belief items \cite{kosinski2024evaluating,strachan2024testing}. Success is brittle in ways a genuine representation should not be: trivial alterations that leave the logical structure intact collapse performance \cite{ullman2023trivial}, and stress tests suggest much apparent competence reflects surface regularities in the item format \cite{shapira2024clever,sap2022neural}, which has pushed benchmarks toward conversational and procedurally generated settings \cite{kim2023fantom,gandhi2023understanding}. Social simulation meets the same question from the other side. Generative agents sustain believable individual and collective behavior \cite{park2023generative,park2022social} and reproduce aggregate patterns from human subject studies \cite{argyle2023one,aher2023using,zhou2024sotopia}, yet they flatten into caricature when asked to represent identity groups \cite{cheng2023compost} and misstate the opinion distributions they stand for \cite{santurkar2023whose}. Those critiques concern whom a simulation represents. We ask a mechanism level question that survives even when representativeness is not at issue: within a single dyad, does the agent maintain a partner model at all, or does it generate plausible continuations of a conversation? Because verbal reports can diverge from the process that produced the behavior \cite{nisbett1977telling}, we place the inference inside a live interaction the model itself produced and score it against a hidden state with consequences, rather than against coherence.

\paragraph{Negotiation as an instrument.}
Negotiation has long been used in psychology and economics precisely because it makes hidden preferences consequential \cite{raiffa1982art,bazerman2000negotiation}. The incompatibility bias, in which negotiators assume opposed interests on issues where they actually agree, is a well replicated human failure that depends on a faulty partner model rather than on faulty arithmetic \cite{thompson1990social,thompson1991information}. Language model negotiation has been studied for capability, through learned dialogue agents \cite{lewis2017deal,he2018decoupling} and multi agent platforms that score deal quality and strategic behavior \cite{abdelnabi2024cooperation,bianchi2024negotiationarena}. We differ in instrumentation rather than in task: we enumerate the full Pareto frontier so that efficiency is exact rather than relative, and we pair each negotiation with counterfactual probes of the partner model, which lets us connect an economic failure to a specific representational one. This follows a broader move toward evaluations whose targets are checkable by construction rather than by judgment \cite{he2025matpbench}.

\paragraph{Roles and mental state modeling in multi-agent systems.}
Multi agent architectures assign agents to roles, and role assignment changes what information an agent holds. Role structured systems have been built for judicial debate \cite{he2024agentscourt}, collaborative causal explanation \cite{he2023lego}, and clinical Socratic tutoring \cite{he2026clintutor}, and work on strategic persuasion makes the partner model explicit by planning rebuttals against a reviewer's inferred mental state \cite{he2026rebuttalagent}. Multi agent deliberation is also used to improve reasoning \cite{du2023debate} and to arbitrate between candidate answers \cite{zheng2023judging,yang2026marssql}. All of these designs presume that an agent can represent what another party wants or knows. Our audit tests that presumption directly, and finds the representation substantially weaker, and substantially more egocentric, than the interaction transcript implies.

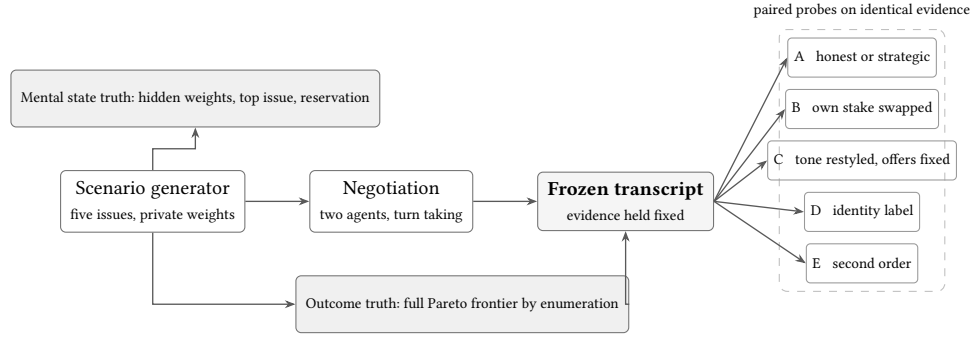
\begin{figure*}[t]
\centering
\begin{tikzpicture}[
  scale=0.94, transform shape,
  font=\small,
  box/.style={draw=black!55, rounded corners=2pt, align=center, inner sep=4pt, minimum height=8mm},
  truth/.style={box, fill=black!6, align=center},
  arm/.style={draw=black!40, rounded corners=1.5pt, align=center, inner sep=2.5pt, font=\scriptsize, minimum height=5.5mm},
  flow/.style={-{Stealth[length=4pt]}, black!65, line width=0.5pt}
]
\node[box] (gen) {Scenario generator\\{\scriptsize five issues, private weights}};
\node[box, right=9mm of gen] (neg) {Negotiation\\{\scriptsize two agents, turn taking}};
\node[box, right=9mm of neg, fill=black!4] (frozen) {\textbf{Frozen transcript}\\{\scriptsize evidence held fixed}};

\node[truth, above=6mm of gen, xshift=6mm] (mind) {{\scriptsize Mental state truth: hidden weights, top issue, reservation}};
\node[truth, below=6mm of neg, xshift=10mm] (out) {{\scriptsize Outcome truth: full Pareto frontier by enumeration}};

\node[arm, right=10mm of frozen, yshift=13mm] (b) {B \; own stake swapped};
\node[arm, below=1.6mm of b] (c) {C \; tone restyled, offers fixed};
\node[arm, below=1.6mm of c] (d) {D \; identity label};
\node[arm, below=1.6mm of d] (e) {E \; second order};
\node[arm, above=1.6mm of b] (a) {A \; honest or strategic};

\draw[flow] (gen) -- (neg);
\draw[flow] (neg) -- (frozen);
\draw[flow] (gen.north) -- ++(0,3mm) -| (mind.south);
\draw[flow] (gen.south) -- ++(0,-3mm) |- (out.west);
\draw[flow] (out.east) -| ($(frozen.south)+(0,-3mm)$) -- (frozen.south);
\foreach \t in {a,b,c,d,e} { \draw[flow] (frozen.east) -- (\t.west); }

\begin{scope}[on background layer]
\node[draw=black!25, dashed, rounded corners=3pt, fit=(a)(e), inner sep=3pt] (probes) {};
\end{scope}
\node[font=\scriptsize, above=0.5mm of probes] {paired probes on identical evidence};
\end{tikzpicture}
\caption{The audit. A generator fixes both ground truths before any agent speaks: the partner's hidden weight vector and the complete Pareto frontier. Two agents negotiate once, and the transcript is then frozen. Intervention A is carried by the negotiation itself. Interventions B through E are counterfactual probes applied to the same frozen record, each changing exactly one factor while the evidence about the partner stays byte identical.}
\label{fig:protocol}
\end{figure*}

\section{Audit Design}
\label{sec:design}

\subsection{Why negotiation}

We need a social simulation in which a researcher can check, from outside, whether an agent recovered something that was genuinely hidden from it. Negotiation over several issues gives this for free. Each side is handed a private table of points, so the partner's preferences are hidden by construction rather than by stipulation, and the value of every possible agreement is computable, so the quality of the interaction is not a matter of judgment. Crucially the two things we want to relate come from the same object: the partner's weight vector both is the thing to be inferred and determines which agreements are efficient.

\subsection{Scenario construction}

Each scenario has five issues, each resolvable at one of five levels. The two sides hold separate point tables, each summing to 100, and every weight within a side is distinct so that both the top priority and the full priority ranking are well defined inference targets. Issues fall into five types. Two integrative issues are weighted asymmetrically, so trading across them creates joint value. One distributive issue is weighted equally and opposed, so it is close to zero sum. One compatible issue points the same direction for both sides, so its mutual optimum is available to anyone who notices. One filler issue carries little weight for either side. Reservation values are set at 30 percent of each side's maximum, and we verify that the zone of possible agreement is non empty in every scenario.

We enumerate all $5^5$ complete packages and sweep out the exact Pareto frontier, the joint maximum, and each side's individual maximum. A self test over 200 freshly generated scenarios asserts every structural property we rely on before any scenario enters the study; Appendix~\ref{app:scenarios} lists the assertions. The released scenario file contains \NScenarios\ scenarios across two framings, an employment offer and a supplier contract.

\subsection{Negotiation and freezing}

Each dyad is a turn taking negotiation between two model backed agents. An agent sees the public issue sheet, its own private point table, its reservation value, and the transcript so far, and replies with a message, a complete proposed package, and an acceptance decision in a fixed structured format. An agreement closes when one side accepts the other's standing proposal. We cross two factors in the negotiation itself. Model assignment is swapped between the two roles so that any finding is not an artifact of which family occupies which seat. Disclosure is either honest, where an agent is told to describe its priorities truthfully, or strategic, where it is explicitly permitted to misrepresent them. Every resulting transcript is then frozen. No probe changes the negotiation, and no negotiation is rerun after a probe.

\subsection{Five paired interventions}

Freezing is what makes the audit clean. The transcript is the entire evidence base about the partner, so if we hold it fixed and move one other factor, any change in the reported partner model cannot be attributed to the agent having learned something new. Figure~\ref{fig:protocol} shows the full protocol. The nine arms are participant\_base and analyst\_base, the mirror and inverse arms of B, the assertive and accommodating arms of C, the senior and junior arms of D, and second\_order for E; Table~\ref{tab:arms-results} reports them all. Intervention A is carried by the negotiation itself, since honest and strategic dyads produce different transcripts. Interventions B through E are applied post hoc to the same frozen record.

Intervention B is the projection test, and it is the reason we include an analyst framing. Swapping a participant's own point table would leave the transcript inconsistent with the moves that participant made, and a model might reasonably react to the inconsistency. Instead we give the same transcript to an uninvolved reader who holds a stake of their own. The evidence about the partner is then literally identical across the mirror and inverse arms, and only the reader's private interest differs. A perfect inferential reader would report the same partner model in both, so any divergence is egocentric projection in the sense used in the perspective taking literature \cite{keysar2003limits,epley2004perspective,nickerson1999how,camerer1989curse}. The analyst base arm controls for the framing itself.

Intervention C restyles the partner's messages into an assertive or an accommodating register. Proposals are stored structurally and re-rendered by the harness, so the offer sequence a probe sees is identical across the two arms by construction and only the surrounding wording changes; we additionally verify that the multiset of numerals in the restyled messages matches the original. Intervention D attaches a seniority description to the partner while holding every message fixed. We deliberately use experience level rather than a protected demographic attribute: the question is whether an unrelated status cue displaces evidence, and seniority answers it without constructing demographic stereotypes.

Intervention E reverses the direction of recursion and asks the agent what its partner believes about it. This arm admits two scoring targets, and separating them is informative. Scored against the agent's own true top priority, a hit means the agent assumed the partner could see it, which is an assumption of transparency rather than an inference. Scored against what the partner actually reported in its own base probe, a hit means the agent modeled the partner's belief, including when that belief is mistaken. We report both.

\subsection{Measures}

Every probe presents the public issue sheet, the reader's private point table, and the frozen transcript, and requests a structured report of the partner's priorities, compatible issues, and reservation value; Appendix~\ref{app:probes} gives the elicitation in full. Partner model accuracy has four components: top priority accuracy, the Kendall correlation between the reported and true priority rankings, whether the compatible issue is identified as compatible, and the absolute error in the estimated reservation value. We also report the normalized distance between the reported and true weight vectors. Social outcome quality has four components: the agreement rate, whether the deal is Pareto optimal, the fraction of attainable joint value destroyed, and whether the deal captures the compatible issue's mutual optimum. All interval estimates are 95\% bootstrap intervals over 2000 resamples at the dyad level, paired wherever the two arms read the same unit, and are not corrected for multiple comparisons.

No model is trained or fine tuned at any point. Scenario generation is seeded, negotiation and probe runs are resumable, and every number in this paper is emitted by the analysis script into a macro file the paper includes, so no figure in the text is transcribed by hand. Two verification gates are released with the code, one over the source and the frozen records and one over the compiled manuscript; Appendix~\ref{app:repro} gives the protocol.

\section{Results}
\label{sec:results}

\subsection{Fluent agreement, inefficient deals}

The negotiations succeed as conversations and fail as agreements. Across \NDyads\ dyads the agents reach agreement \AgreementRate\% of the time in a mean of \MeanTurns\ turns, with \ParseFailures\ protocol failures. Read as transcripts, these are competent negotiations: the agents exchange packages, make concessions, and close.

\begin{figure*}[t]
\centering
\includegraphics[width=\textwidth]{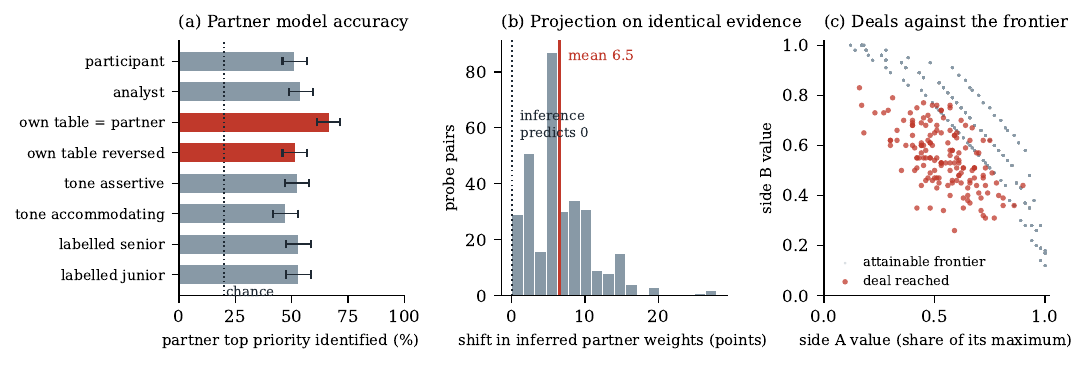}
\caption{(a) Partner top priority accuracy by probe arm, with 95\% bootstrap intervals and the uniform baseline marked. The two highlighted arms differ only in the reader's own payoff table. (b) Distribution of the shift in the inferred partner weight vector between those two arms. The evidence is identical across the pair, so inference predicts zero. (c) Every agreement plotted against the attainable frontier of its own scenario, with both axes normalized by each side's maximum. Deals sit consistently inside the frontier rather than on it.}
\label{fig:results}
\end{figure*}

Scored against the frontier, they are not, as Figure~\ref{fig:results}(c) shows. Only \ParetoRate\% of agreements are Pareto optimal, and the average deal destroys \EfficiencyLoss\% of the joint value that was available. The most diagnostic number concerns the compatible issue, where the two sides want the same level, so capturing it costs nothing and requires only noticing. It is missed in \CompatMissRate\% of deals. This is the signature of the incompatibility bias \cite{thompson1990social}, reproduced by agents that never had to overcome a competitive emotion, only a representational gap. Disclosure moves it little: efficiency loss is \EffHonest\% under honest and \EffStrategic\% under strategic disclosure, a difference of \HonEffMean\ percentage points with interval \HonEffCI, so the shortfall is not mainly a product of deception.

These numbers need a reference point and the design supplies an exact one: average over every package both sides would accept. Such a package is Pareto optimal \BaseParetoRate\% of the time, destroys \BaseEfficiencyLoss\% of joint value, and captures the compatible optimum \BaseCompatHitRate\% of the time. Comparing each agreement with the baseline of its own scenario, agent minus baseline throughout, the agents do recover value: joint value destroyed differs by \EffGapMean\ points, interval \EffGapCI. On the other two the difference is indistinguishable from zero: \ParetoGapMean\ points on Pareto optimality, interval \ParetoGapCI, and \CompatGapMean\ points on the compatible issue, interval \CompatGapCI. On exactly the measures that require reading the structure of the partner's interests, these negotiations are indistinguishable from a random acceptable package.

\subsection{Partner models are weak}

Table~\ref{tab:arms-results} and Figure~\ref{fig:results}(a) report partner model accuracy by arm. In the base arm the negotiator identifies the partner's top priority \TopParticipantBase\% of the time against a uniform baseline of 20\%, recovers the priority ranking at a Kendall correlation of \TauParticipantBase, identifies the compatible issue \CmpParticipantBase\% of the time, and misestimates the partner's reservation value by \ResParticipantBase\ points on a 100 point scale. The agents are therefore above chance and far from accurate, and the reservation error is large enough that an agent negotiating against its own estimate would routinely misjudge whether a deal was even feasible. Stated confidence does not flag the misses: over \ConfGapN\ first order probes it spans only \ConfMin\ to \ConfMax\ and averages \ConfCorrect\ when the top priority is right against \ConfIncorrect\ when it is wrong, a gap of \ConfGapMean\ with interval \ConfGapCI.

\begin{table}[t]
\caption{Partner model accuracy by probe arm. Top is top priority accuracy in percent, $\tau$ is the rank correlation with the true priority ordering, Cmp is compatible issue identification in percent, and Res is absolute reservation error in points.}
\label{tab:arms-results}
\small
\begin{tabular}{@{}lrrrr@{}}
\toprule
Arm & Top & $\tau$ & Cmp & Res \\
\midrule
participant\_base & \TopParticipantBase & \TauParticipantBase & \CmpParticipantBase & \ResParticipantBase \\
analyst\_base & \TopAnalystBase & \TauAnalystBase & \CmpAnalystBase & \ResAnalystBase \\
\midrule
analyst\_mirror & \TopAnalystMirror & \TauAnalystMirror & \CmpAnalystMirror & \ResAnalystMirror \\
analyst\_inverse & \TopAnalystInverse & \TauAnalystInverse & \CmpAnalystInverse & \ResAnalystInverse \\
\midrule
tone\_assertive & \TopToneAssertive & \TauToneAssertive & \CmpToneAssertive & \ResToneAssertive \\
tone\_accommodating & \TopToneAccommodating & \TauToneAccommodating & \CmpToneAccommodating & \ResToneAccommodating \\
\midrule
label\_senior & \TopLabelSenior & \TauLabelSenior & \CmpLabelSenior & \ResLabelSenior \\
label\_junior & \TopLabelJunior & \TauLabelJunior & \CmpLabelJunior & \ResLabelJunior \\
\bottomrule
\end{tabular}
\end{table}

The participant and analyst base arms are statistically indistinguishable, with a top priority difference of \FrameTopMean\ percentage points and interval \FrameTopCI. Having negotiated the deal oneself confers no advantage over reading someone else's transcript. This is itself a result about what the partner model is built from: if participation added private information, such as a felt sense of where the partner resisted, the participant arm should lead.

\subsection{Projection displaces inference}

The projection test is the clearest effect in the study. The mirror and inverse arms show the same reader the same transcript and differ only in the reader's own private point table. A reader performing inference would produce identical answers. Figure~\ref{fig:results}(b) shows the result. Instead the reported partner weight vector moves by \ProjShift\ points, with interval \ProjShiftCI, the reported top priority moves by \ProjTopMean\ percentage points with interval \ProjTopCI, and the reported ranking moves by \ProjTauMean\ in rank correlation with interval \ProjTauCI. None of these intervals include zero.

The direction is as informative as the magnitude. Accuracy is higher in the mirror arm, \TopAnalystMirror\%, than in the inverse arm, \TopAnalystInverse\%. In the mirror arm the reader's own table has been set equal to the partner's weights, so projecting the self onto the partner returns the correct answer without any inference having occurred. Apparent theory of mind competence therefore rises and falls with how similar the reader happens to be to the partner, which is a warning about benchmark construction as much as about agents.

\subsection{Surface cues move the reading far less than the self does}

Interventions C and D are small where intervention B is large, and the contrast is the point. Restyling the partner's messages between an assertive and an accommodating register, while holding the offer sequence identical, changes top priority accuracy by \ToneTopMean\ percentage points with interval \ToneTopCI\ and rank correlation by \ToneTauMean\ with interval \ToneTauCI. The tone effect is therefore real but modest: it is roughly a third of the \ProjTopMean\ point swing produced by changing nothing except the reader's own payoff sheet, and it runs in the direction of the assertive register, \TopToneAssertive\% against \TopToneAccommodating\%, which suggests that a firmer restatement of the same position reads as a clearer signal of priority. Attaching a senior or junior label to the partner changes top priority accuracy by \IdTopMean\ percentage points with interval \IdTopCI\ and rank correlation by \IdTauMean\ with interval \IdTauCI, so the priority structure the agent reports is essentially unmoved by who it is told the partner is.

The two manipulations separate cleanly on the walk away estimate. Under the senior label the reported reservation value rises by \IdResLevelMean\ points with interval \IdResLevelCI, and since the truth is unchanged the absolute error rises with it by \IdResMean\ points with interval \IdResCI. Tone moves neither, at \ToneResLevelMean\ points with interval \ToneResLevelCI. An agent told that its counterpart is senior therefore credits that counterpart with a higher walk away point on the same transcript and the same offers, although the label carries no information about the hidden reservation value at all. Status moves the agent's guess about how hard the partner will push but not its model of what the partner wants, and register nudges how legible a stated priority appears without changing what the agent takes that priority to be. Taken with the projection result, this sharpens the diagnosis. The failure is not general suggestibility, since both surface manipulations move the reported partner model by far less than the reader's own table does. It is specific: where evidence underdetermines the answer, the agent falls back on itself.

\subsection{Tracking the message, not the mind}

Intervention E gives the sharpest statement of the paper's question. Asked what its partner believes about it, an agent matches the partner's actually reported belief \SecondOrderHit\% of the time. That same partner's belief is correct only \PartnerActuallyRight\% of the time. The agent is therefore a better model of its partner's stated position than its partner is of the hidden truth, and it remains accurate about the partner's belief precisely when that belief is wrong.

Scored instead against the agent's own true top priority, the second order answer hits \AssumedTransparency\% of the time, a difference of \TransGapMean\ percentage points with interval \TransGapCI. Both targets are reached more often than first order accuracy about the hidden weights. The most economical account is that the agents maintain a shared representation of the conversation, mutually accessible and readable off the transcript by both parties, and that both first and second order answers are drawn from it. What they do not maintain is an independent model of the private state the conversation was supposed to be evidence about. The simulation is coherent at the level of the message and thin at the level of the mind.

\subsection{Partner model quality and deal quality}

Finally we connect the two ground truths. Splitting dyads at the median partner model accuracy, deals in the more accurate half destroy \LinkEffHigh\% of available value against \LinkEffLow\% in the less accurate half, and capture the compatible issue \LinkCompatHigh\% of the time against \LinkCompatLow\%. The direction favors better partner models on the compatible issue, the component most directly dependent on recognizing what the other side wants, while overall efficiency is close to flat. We report this as a descriptive association across \LinkN\ dyads rather than a causal claim, since partner model accuracy is measured after the negotiation rather than manipulated before it.

\section{Discussion}
\label{sec:discussion}

\paragraph{Surface plausibility is not a validity check.}
The transcripts in this study would pass any reading based inspection. The agents are polite, responsive, and cooperative, they converge, and they almost never fail to reach a deal. Every failure we report is invisible from the transcript and visible only against ground truth the transcript does not contain. A simulation evaluated by reading it, or by asking a language model whether it looks realistic, cannot distinguish an agent that models its partner from one that continues a conversation. If a study's conclusion depends on the agents having represented each other, it needs a latent variable the agents did not see and the researcher can score.

\paragraph{Projection is the specific failure mode.}
Our results do not support a blanket claim that agents lack partner models. They are above chance on every mental state measure, and tone and status cues move their inferred priorities by far less than their own stake does. What they do is fall back on their own preferences when the evidence underdetermines the answer, strongly enough to move the reported partner model when nothing about the partner has changed. Interventions aimed at improving simulated social reasoning should therefore target the self anchor rather than general reasoning capacity, which improves along an axis of its own \cite{zhu2026loopedlm}: making the agent's own stake less salient at inference time, or forcing an explicit hypothesis about the partner before the agent's own table is revealed, are both testable mitigations that follow from the mirror and inverse comparison. The same comparison warns about benchmark construction: because mirror readers are more accurate than inverse readers without doing more inference, any evaluation whose characters have similar preferences will overstate theory of mind competence. We recommend that partner modeling evaluations report accuracy separately for aligned and opposed preference pairs, which our design makes routine.

\paragraph{Implications for deployed multi-agent systems.}
Multi agent architectures increasingly assign agents to negotiate, mediate, or advocate on a user's behalf, and role structured systems presume that an agent can represent another party's interests \cite{he2024agentscourt,he2023lego,he2026clintutor,he2026rebuttalagent}. Our efficiency numbers suggest a specific risk profile. The failure is not that agreements fail to be reached. Agreements are reached almost always, which means the failure is silent: value is lost on exactly the issues where both parties could have won, and the loss would not be detectable by a user reading the transcript or by a satisfaction rating collected afterward. Systems of this kind should be instrumented with outcome measures computed against the space of possible agreements, not with acceptance rates.

\section{Limitations}
\label{sec:limitations}

Our scenarios are synthetic and structurally regular. Real negotiations have unequal information, outside options that change during the interaction, relationships that persist beyond the deal, and issues that resist numerical weighting. That regularity is what buys us exact ground truth, and we accept it as a deliberate trade; how far synthetic material of this kind carries is an open question in its own right \cite{qin2025scalingsynth}.

We study two model families in one interaction structure, and all \NProbes\ probes are answered by a single reader model. Projection appears in each of the \ProjTopCells\ cells of model assignment and seat, between \ProjTopCellMin\ and \ProjTopCellMax\ points, so it does not depend on which family wrote the transcript; still, every mental state result is one reader's, and a second reader is the obvious replication. We cannot claim the results generalize to all agents, to longer horizons, where generation itself becomes less stable \cite{he2026stablelongform}, or to groups larger than a dyad. We also elicit partner models by asking for them, which is a verbal report and may diverge from whatever representation drove the agent's moves \cite{nisbett1977telling}. This is why the outcome measures matter independently: the efficiency shortfall is a behavioral fact that does not depend on the probes being faithful. The link between partner model accuracy and deal quality is likewise descriptive, since accuracy is measured after the negotiation closes; manipulating partner model quality beforehand, for instance by varying how much of the partner's structure is disclosed, would test the causal claim and is the natural next step.

Finally, our identity intervention uses seniority rather than demographic attributes, a deliberate choice to avoid constructing stereotype probes about protected groups. Our near null result on identity priorities should therefore not be read as evidence that demographic labels would also be ignored; that question needs a different and more careful study design.

\section{Conclusion}
\label{sec:conclusion}

We built a social simulation in which the partner's mind and the quality of the social outcome both have exact answers, ran \NDyads\ negotiations, froze every transcript, and applied \NProbes\ counterfactual probes that changed one factor at a time on identical evidence. The agents were fluent and cooperative and reached agreement almost always, while destroying \EfficiencyLoss\% of available joint value and missing the issue of perfectly aligned interest in \CompatMissRate\% of deals. Probing the frozen record located the gap: inferred partner preferences move with the reader's own stake when nothing about the partner has changed, a tone rewrite and a status label move them far less, and agents predict their partner's stated beliefs more accurately than they recover the hidden state those beliefs are about. The simulations track the message well and the mind poorly, and measuring that only requires building the ground truth in before the conversation starts.

\bibliographystyle{ACM-Reference-Format}
\bibliography{references}

\appendix

\section{Scenario Generator Properties}
\label{app:scenarios}

Every generated scenario is checked against a fixed list of structural assertions before it enters the study. Weights within each side sum to 100 and are pairwise distinct, so the top priority and the complete ranking are both well defined. Each of the five issue types appears exactly once. The compatible issue ramps in the same direction for both sides and therefore admits a mutual optimum. The distributive issue carries equal weight for the two sides and ramps in opposite directions. The Pareto frontier is non empty, the joint maximum lies on it, and the zone of possible agreement under both reservation values is non empty. The two sides differ in top priority, which is required for the inference target to be non trivial. The scoring function is verified to round trip on randomly drawn packages. These assertions run over 200 freshly generated scenarios as part of the verification gate, and any failure stops the study before a single agent is called.

\section{Probe Elicitation}
\label{app:probes}

Each probe presents the public issue sheet, the reader's private point table, and the frozen transcript, and requests a structured report containing the partner's top priority, a complete priority ranking, a weight vector over the five issues, any issues believed to be compatible, an estimated reservation value, and a confidence rating. The confidence rating enters no accuracy measure and is reported separately in Section~\ref{sec:results}, since stated confidence and the knowledge boundary it is supposed to mark are known to come apart \cite{he2025mmboundary}. The participant framing addresses the reader as the negotiator. The analyst framing addresses an uninvolved reader with a stake of their own in deals of this kind, which permits varying the reader's table without making the transcript inconsistent with the moves it records.

Tone arms are produced by a rewriting pass that is instructed to preserve every substantive claim, issue mention, numeral, and position, and to change only register. Proposals are never routed through the rewriter: they are stored as structured packages and re-rendered identically in every arm, so the offer sequence a probe sees is fixed by construction. We also compare the multiset of numerals in the restyled messages against the original, record the result with each probe, and fail the gate if the mismatch rate exceeds a fixed threshold.

\section{Reproducibility}
\label{app:repro}

No model is trained or fine tuned. The study is scenario generation, one pass of negotiation, and one pass of probes over the frozen transcripts. Scenario generation is seeded and deterministic. Negotiation and probe runs are resumable and skip any record already present, so a partial run can be completed without duplicating calls. Every number reported in this paper is generated by the analysis script into a macro file that the paper includes, so no figure in the text is transcribed by hand.

\end{document}